\documentclass{article}

\usepackage{microtype}
\usepackage{graphicx}
\usepackage{subfigure}
\usepackage{booktabs} % for professional tables

\usepackage{hyperref}

\usepackage[accepted]{icml2025}

\usepackage{amsmath}
\usepackage{amssymb}
\usepackage{mathtools}
\usepackage{amsthm}

\usepackage{cuted}
\usepackage{multirow}
\usepackage{subcaption}

\usepackage[capitalize,noabbrev]{cleveref}

\theoremstyle{plain}

\theoremstyle{definition}

\theoremstyle{remark}

\usepackage[textsize=tiny]{todonotes}

\icmltitlerunning{MakeAnything: Harnessing Diffusion Transformers for Multi-Domain Procedural Sequence Generation}

\begin{document}
\twocolumn[{%
\renewcommand\twocolumn[1][]{#1}%

% \twocolumn[
\icmltitle{MakeAnything: Harnessing Diffusion Transformers for Multi-Domain Procedural Sequence Generation}





\begin{icmlauthorlist}
\icmlauthor{Yiren Song}{yyy}
\icmlauthor{Cheng Liu}{yyy}
\icmlauthor{Mike Zheng Shou}{yyy}
\end{icmlauthorlist}

\icmlaffiliation{yyy}{Show Lab, National University of Singapore, Singapore}

\icmlcorrespondingauthor{Mike Zheng Shou}{mike.zheng.shou@gmail.com}

% \author{
% Yiren Song\thanks{Equal contribution.} \quad Xiaokang Liu\footnotemark[1] \quad Mike Zheng Shou\thanks{Corresponding author.} \\
% Show Lab, National University of Singapore \\
% % \texttt{yiren@nus.edu.sg, xliu@u.nus.edu, mike.zheng.shou@gmail.com}
% }

\icmlkeywords{Machine Learning, ICML}

\begin{center}
    \centering
    \captionsetup{type=figure}
    \includegraphics[width=.99\textwidth]{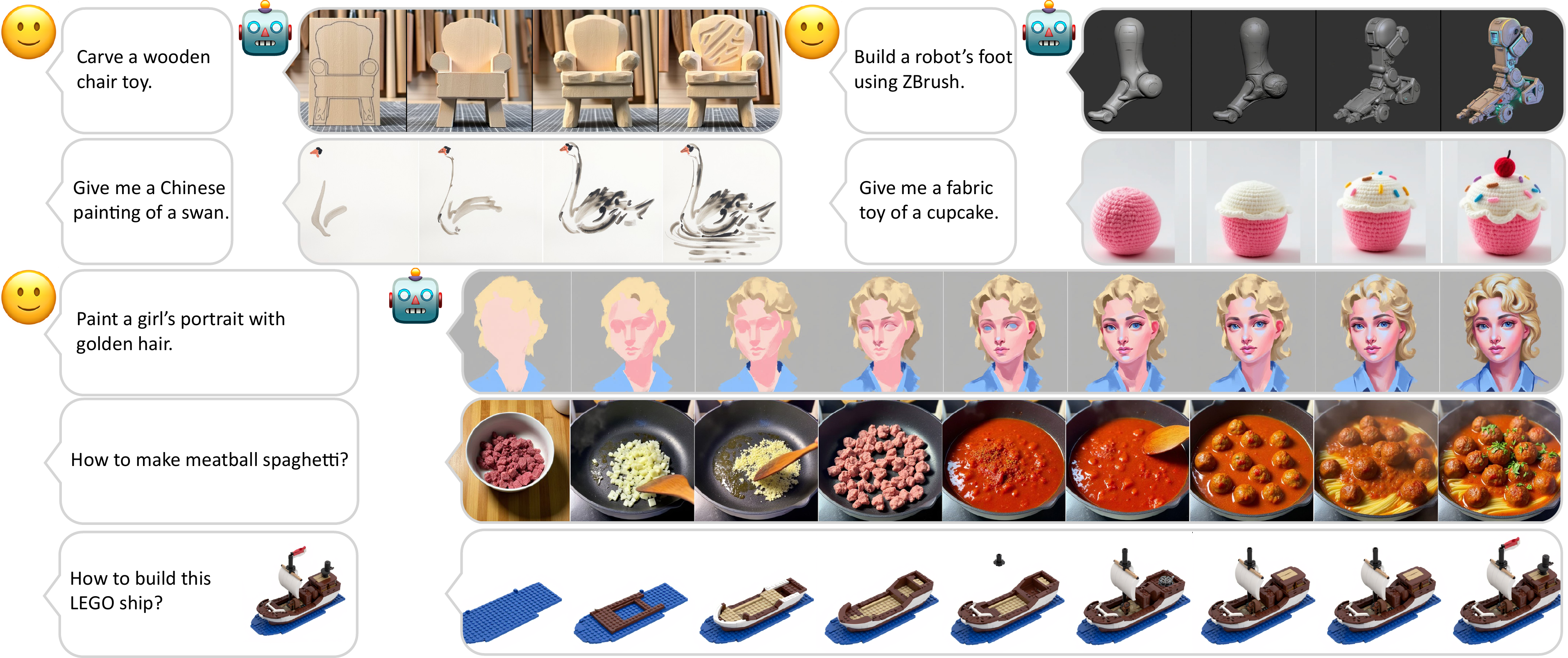}
    \captionof{figure}{We introduce MakeAnything, a tool that realistically and logically generates step-by-step procedural tutorial  for activities such as painting, crafting, and cooking, based on text descriptions or conditioned images.}
\end{center}%
}]



\printAffiliationsAndNotice{}  % leave blank if no need to mention equal contribution


\begin{abstract}
A hallmark of human intelligence is the ability to create complex artifacts through structured multi-step processes. Generating procedural tutorials with AI is a longstanding but challenging goal, facing three key obstacles: (1) scarcity of multi-task procedural datasets, (2) maintaining logical continuity and visual consistency between steps, and (3) generalizing across multiple domains. To address these challenges, we propose a multi-domain dataset covering 21 tasks with over 24,000 procedural sequences. Building upon this foundation, we introduce MakeAnything, a framework based on the diffusion transformer (DIT), which leverages fine-tuning to activate the in-context capabilities of DIT for generating consistent procedural sequences. We introduce asymmetric low-rank adaptation (LoRA) for image generation, which balances generalization capabilities and task-specific performance by freezing encoder parameters while adaptively tuning decoder layers. Additionally, our ReCraft model enables image-to-process generation through spatiotemporal consistency constraints, allowing static images to be decomposed into plausible creation sequences.  Extensive experiments demonstrate that MakeAnything surpasses existing methods, setting new performance benchmarks for procedural generation tasks. Code is released at \href{https://github.com/showlab/MakeAnything}{https://github.com/showlab/MakeAnything}
\end{abstract}

% 人类智能的一个标志是能够通过结构化的多步骤过程创造复杂的工艺品。让AI生成连贯的创作教程是一直想要但是困难的，面临三个挑战：（1）多任务过程数据集的稀缺（2）在步骤之间保持逻辑连续性和视觉一致性，以及 （3）多个领域的泛化。我们收集了多任务数据集，涵盖了21个任务，共计超过24,000个序列。提出了一个基于扩散变换器（DIT）的框架MakeAnything，通过微调激活DiT的in-context 能力来生成前后一致的过程序列。我们为图像生成引入了非对称低秩适应（LoRA），通过冻结编码器参数同时适应性调整解码器层在泛化能力与任务特定性能之间实现平衡。此外，我们的ReCraft模型通过时空一致性约束实现了从图像到过程的生成，允许将静态图像分解为合理的创造序列。广泛的实验表明，MakeAnything 超越了现有方法，为流程生成任务设立了新的性能基准。

% 人类智能的一个标志是能够以逐步方式创造复杂的人工制品。然而，为绘画、手工制作和烹饪等任务生成多步骤的流程仍然充满挑战。我们提出了 **MakeAnything**，一个利用扩散变换器（Diffusion Transformer, DIT）生成高质量指令序列的框架。我们精心构建的数据集涵盖了绘画、手工制作、SVG创建、乐高搭建和烹饪等21个类别，共计超过15,000个序列。为减轻在小规模类别上的过拟合问题，我们采用了非对称LoRA策略，在泛化能力与任务特定性能之间实现平衡。此外，我们引入了 **ReCraft**，支持将现有图像分解为一致的逐步序列，以及 **Strategy Adapter**，通过少量样本快速适应未见任务的需求。广泛的实验表明，MakeAnything 超越了现有方法，为流程生成任务设立了新的性能基准。

% % 图
%1. 又大又好看的teaser 概括任务，text2， image2， 策略encoder？
%2. method illustrator （simple）(可选）
%3. method illustrator （detail）
%4. dataset 统计图，类型和比重, 合成数据方案
%5. 结果图，text2， image2，策略encoder, 拼大图
%6. 类内泛化，类间泛化， real world application
%7. 对比实验结果, with PP
%8. 消融实验, 1. 不对称LORA 2. 

% 表
% T1， 对比实验定量结果
% T2， 我们方法的评估结果， FID？， GPT4 score
% T3， User study结果

% 文字
% 1. 方法细节， 公式化描述，  策略 encoder？ 讲清楚最关键的不对称LORA和 泛化。 和IC LORA区别！！位置编码！！ 4帧数据LORA 如何出9帧
% 2. Relate work 缩减篇幅
% 3. 讨论实验结果，过程策略的异同，雕刻减法策略， 绘画加法策略， 转变过程（变形金刚，烹饪）same prompt, different B
% 4. Userstudy 细节
% 5. 图和表的分析

%数据集概念的分布不均匀

% motivation， 每个模块的motivation， key insights， finding， 如何组合这些Blora? 不同B加权

% 实验
% 1. Flux LoRA Merge, replace ReCraft baseModel
% 2. Strategy Adapter training
% 3. VAE pintu, S dataset

% Todo：1.teaser 改caption 2.Adapter 3 图3大小写 4.图表分析 5.检查引用    
\section{Introduction}

A defining characteristic of human intelligence—and a key differentiator from other species—is the capacity to create complex artifacts through structured step-by-step processes. In computer vision, generating such procedural sequences for tasks like painting, crafting, product design, and culinary arts remains a significant challenge. The core difficulty lies in producing multi-step sequences that maintain logical continuity and visual consistency, requiring models to both capture intricate visual features and understand causal relationships between steps. This challenge becomes particularly pronounced when handling diverse domains and styles without compromising generation quality—a problem space that remains underexplored.

Existing research primarily focuses on decomposing painting processes, with early methods employing reinforcement learning/optimization algorithms through stroke-based rendering to approximate target images. Subsequent works like ProcessPainter \cite{processpainter} and PaintsUndo \cite{paintsundo} utilize temporal models on synthetic datasets, while Inverse Painting \cite{inverse}redicts the order of human painting, generating the painting process by region. However, these approaches remain limited to single-task scenarios and exhibit poor cross-domain generalization. Furthermore, ProcessPainter's Animatediff-based framework constrains modifications to minor motion adjustments, making it unsuitable for categories requiring structural transformations (e.g., recipes or crafts). Although Diffusion Transformer (DIT) \cite{dit}-based video generation models can produce long sequences, their effectiveness is hindered by distribution shifts in training data when generating complex procedural workflows.

We posit that replicating human creative intelligence requires both high-quality multi-task procedural data and advanced methodology design.. To this end, we curate a comprehensive multi-domain dataset spanning 21 categories (including painting, crafts, SVG design, LEGO assembly, and cooking) with over 24,000 procedurally annotated sequences—the largest such collection for step-by-step creation tasks. Methodologically, we propose MakeAnything, a novel framework that harnesses the in-context capabilities of Diffusion Transformers (DIT) through LoRA fine-tuning to generate high-quality instructional sequences.

Addressing the challenge of severe data scarcity (some categories have as few as 50 data entries.) and imbalanced distributions, we employ an asymmetric low-rank adaptation (LoRA)\cite{asymmetry, lora} strategy for image generation. This approach combines a pretrained encoder on large-scale data with a task-specific fine-tuned decoder, achieving an optimal balance between generalization and domain-specific performance.

 To address practical needs for reverse-engineering creation processes, we develop the ReCraft Model—an efficient controllable generation method that decomposes static images into step-by-step procedural sequences. Building upon the pretrained Flux model with minimal architectural modifications, ReCraft introduces an image-conditioning mechanism where clean latent tokens from the target image (encoded via VAE) guide the denoising of noisy intermediate frames through multi-modal attention. Remarkably, this lightweight adaptation enables efficient training with limited data—ReCraft achieves robust performance with just hundreds or even dozens of process sequences per task. During inference, the model recursively predicts preceding frames over concatenated latent representations, effectively reconstructing the creation history from static artworks. 
 % Extensive experiments demonstrate that our framework outperforms existing methods by significant margins, establishing new state-of-the-art benchmarks for procedural generation.

In summary, our contributions are as follows:  
\begin{enumerate}  
\item \textbf{Unified Procedural Generation Framework}: We introduce MakeAnything, the first DIT-based architecture enabling cross-domain procedural sequence synthesis, supporting both text-to-process and image-to-process generation paradigms.
\item \textbf{Technical Innovations}: We employ an asymmetric LoRA architecture for cross-domain generalization and the ReCraft Model for image-conditioned process reconstruction with limited training data.
\item \textbf{Dataset Contribution}: We propose a multi-domain procedural dataset (21 categories, 24K+ sequences) with hierarchical annotations, significantly advancing research in procedural understanding and generation. 
\end{enumerate}

\vspace{-0.5em}
\section{Related Work}

\subsection{Diffusion Models}
Diffusion probability models \citep{ddim,ddpm} are advanced generative models that restore original data from pure Gaussian noise by learning the distribution of noisy data at various levels of noise. Their powerful capability to adapt to complex data distributions has led diffusion models to achieve remarkable success across several domains including image synthesis \citep{rombach2022high,dit}, image editing \citep{instructpix2pix,p2p,stablemakeup,stablehair}, and video gneration \citep{animatediff, svd, processpainter}, evaluation \citep{diffsim}. Stable Diffusion \citep{rombach2022high} (SD), a notable example, utilizes a U-Net architecture and extensively trains on large-scale text-image datasets to iteratively generate images with impressive text-to-image capabilities. The Diffusion Transformer (DiT) model \cite{dit}, employed in architectures like FLUX.1 \cite{flux}, Stable Diffusion 3 \cite{sd3}, and PixArt (pixart), uses a transformer as the denoising network to iteratively refine noisy image tokens. Customized generation methods enable flexible customization of concepts and styles by fine-tuning U-Net \citep{dreambooth} or certain parameters \citep{lora, customdiffusion}, alongside trainable tokens. Training-free customization methods \citep{ipa, ssr, fast} leverage pre-trained CLIP \citep{clip} encoders to extract image features for efficient customized generation. 

% 扩散概率模型/cite{}是先进的生成模型，通过学习各种噪声水平下噪声数据的分布，从纯高斯噪声中恢复原始数据。由于其强大的适应复杂数据分布的能力，扩散模型在多个领域取得了卓越的成就，包括图像合成/cite{}、图像编辑/cite{}、视频生成/cite{}。其中一个著名的例子是稳定扩散模型（Stable Diffusion，SD）\citep{rombach2022high}，该模型采用 U-Net 架构，并在大规模文本图像数据集上进行广泛训练，以迭代方式生成具有令人印象深刻的文本到图像转换能力的图像。扩散变换器（DiT）模型（?），用于如FLUX.1（?）、稳定扩散3（?）和PixArt（?）等架构中，采用变换器作为去噪网络来迭代地细化噪声图像标记。客制化生成方法通过微调U-Net或部分参数，以及使用可训练的token，实现了概念和风格的灵活定制。一些无需训练的客制化方法，如IPA、InstantID、InstantStyle等，利用预训练的CLIP或Arcface编码器提取图像特征，并通过Adapter结构将其注入U-Net的cross-attn层，实现了高效的客制化生成。

\begin{figure*}[htp]
    \centering
    \includegraphics[width=1.0\linewidth]{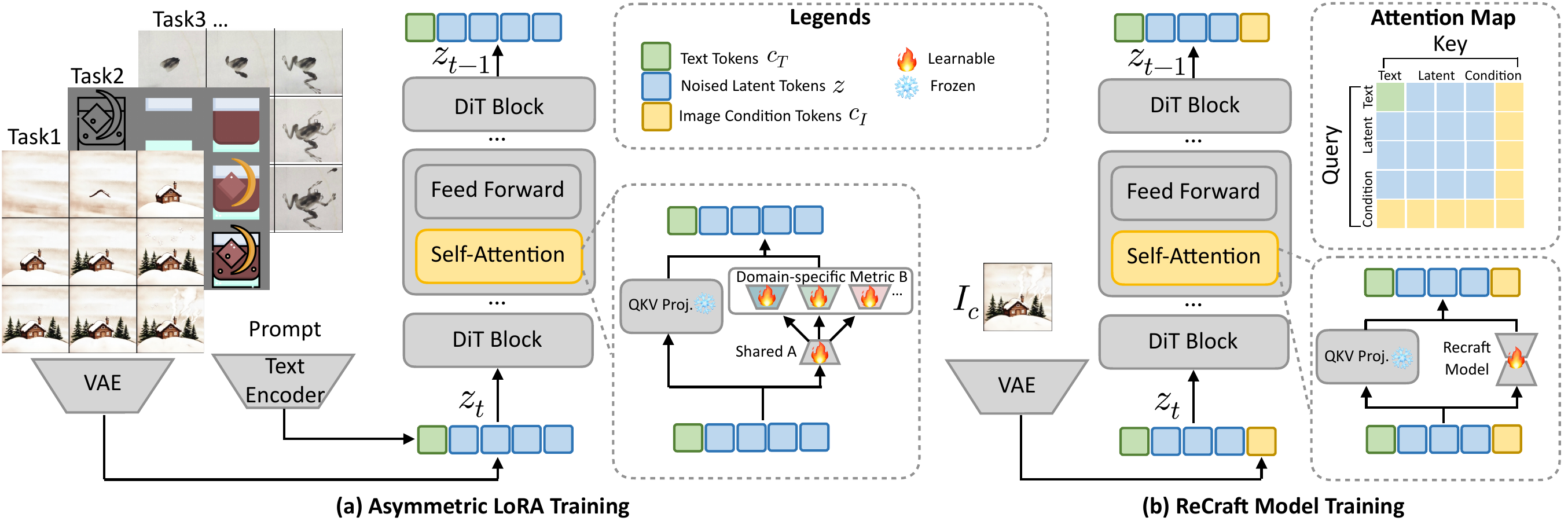}
    \vspace{-6mm}
    \caption{The MakeAnything framework comprises two core components: (1) an Asymmetric LoRA module that generates diverse creation processes from text prompts through asymmetric LoRA, and (2) the ReCraft Model, which constructs an image-conditioned base model by merging pretrained LoRA weights with the Flux foundation model, enabling process prediction via injected visual tokens.}
    \label{fig2}
\end{figure*}
\vspace{-5mm}

\subsection{Controllable Generation in Diffusion Models}

Controllable generation has been extensively studied in the context of diffusion models. Text-to-image models \cite{ddpm, ddim} have established a foundation for conditional generation, while various approaches have been developed to incorporate additional control signals such as images. Notable methods include ControlNet~\cite{controlnet}, enabling spatially aligned control in diffusion models, and T2I-Adapter \cite{t2i}, which improves efficiency with lightweight adapters. UniControl \cite{unicontrol} uses Mixture-of-Experts (MoE) to unify different spatial conditions, further reducing model size.  However, these methods rely on spatially adding condition features to the denoising network’s hidden states, inherently limiting their effectiveness for spatially non-aligned tasks like subject-driven generation. IP-Adapter \cite{ipa} addresses this by introducing cross-attention through an additional encoder. Based on the DiT architecture, OminiControl \cite{ominicontrol} proposes a unified solution that is applicable to both spatially aligned and non-aligned tasks by concatenating condition tokens with noise tokens.

% 可控生成在扩散模型的背景下已被广泛研究。文本到图像模型~\cite{ho2021denoising, ramesh2022hierarchical}为条件生成奠定了基础，同时开发了各种方法以纳入额外的控制信号，如图像。其中值得注意的方法包括 ControlNet~\cite{zhang2023controlnet}，它使扩散模型能够进行空间对齐控制，以及 T2I-Adapter~\cite{mou2023t2i}，该方法通过轻量级适配器提高了效率。然而，这些方法依赖于将条件特征空间地添加到去噪网络的隐藏状态中，这在本质上限制了它们在空间非对齐任务（如主题驱动生成）中的有效性。IP-Adapter~\cite{liu2023ipadapter} 通过引入一个额外的编码器并使用交叉注意力机制来解决这一问题。基于DiT架构， OminiControl提出了一个同时适用于空间对齐和非对齐任务的统一解决方案，通过将condition token和加噪token连接。

% MakeAnything 的结构示意图。由三部分组成，Procedural LoRA， ReCraB Model  和 Strategy Adapter 。 1. 我们首先在MakeAnything dataset上采用LoRA的方式进行预训练， 实现了从文本生成多种制作过程。 2.将第一步的LORA和Flux base model merge作为新的base model，ReCraft Model引入图像条件机制，通过LoRA微调的方式实现了预测参考图的制作过程。 3. 我们在MakeAnything dataset上训练了 Strategy Adapter， 可以从新的制作过程中提取制作策略。

\subsection{Procedural Sequences Generation}

Generating the creation process of paintings or handicrafts is something that has always been desired but is difficult to achieve. The problem of teaching machines "how to paint" has been thoroughly explored within stroke-based rendering (SBR), focusing on recreating non-photorealistic imagery through strategic placement and selection of elements like paint strokes \cite{hertzmann2003survey}. Early SBR methods included greedy searches or required user input \cite{haeberli1990paint,litwinowicz1997processing}, while recent advancements have utilized RNNs and RL to sequentially generate strokes \cite{ha2017neural,zhou2018learning,xie2013artist}. Adversarial training has also been introduced as an effective way to produce non-deterministic sequences \cite{nakano2019neural}. Techniques like Stylized Neural Painting \cite{ kotovenko2021rethinking} have advanced stroke optimization, which can be integrated with neural style transfer. The field of vector graphic generation employs similar techniques \cite{clipdraw, clipvg, cliptexture, clipfont}. However, these methods differ greatly from human creative processes due to variations in artists' styles and subjects. Inverse Painting \cite{inverse} achieves realistic painting process simulation by predicting the painting order and implementing image segmentation. ProcessPainter \cite{processpainter} and Paints Undo \cite{paintsundo} method fine-tunes diffusion models using data from artists' painting processes to learn their true distributions, enabling the generation of painting processes in multiple styles. 

% 生成绘画或手工艺品的创作过程一直是人们所期望但难以实现的。教会机器“如何绘画”的问题已在基于笔触的渲染（SBR）中被彻底探索，这一过程专注于通过策略性放置和选择元素如绘画笔触来重现非真实感图像 \cite{hertzmann2003survey}。早期SBR方法包括贪婪搜索或需要用户输入 \cite{haeberli1990paint,litwinowicz1997processing,hertzmann2022toward, 3Dstroke}，而最近的进展已经利用RNNs和RL来顺序生成笔触 \cite{ha2017neural,zhou2018learning,xie2013artist,singh2022intelli}。对抗训练也被引入作为生成非确定性序列的有效方法 \cite{nakano2019neural}。如《风格化神经绘画》\cite{snp, kotovenko2021rethinking}等技术已推进笔触优化，可以与神经风格转移集成。矢量图形生成领域采用了类似的技术 \cite{clipdraw, clipvg, cliptexture, clipfont}。然而，这些方法因艺术家风格和主题的变化而与人类创造过程大相径庭。Inverse Painting通过预测绘画绘画顺序，和图像分割实现了逼真的绘画过程模拟。  ProcessPainter和Paints Undo方法通过使用艺术家绘画过程的数据对扩散模型进行微调，学习其真实分布，实现了多种风格的绘画过程生成。

\section{Method}
% 3.1节，我们介绍MakeAnything 方法的整体架构，3.2节介绍Procedural LoRA，通过拼图策略和基于DIT的小样本训练实现过程生成 3.3节介绍ReCraft Model， 一种有效的Image condition model 生成和参考图像高度一致的过程序列 3.4节介绍数据收集方法和合成数据管线。

In this section, we begin by exploring the preliminaries on diffusion transformer as detailed in Section 3.1. Next, Section 3.2 introduce the overall architecture of the MakeAnything method. In Section 3.3, we present asymmetric LoRA for Procedual learning. Section 3.4 introduces the ReCraft Model, an effective image condition model that generates procedural sequences highly consistent with reference images. Finally, we introduce the new dataset we proposed in Section 3.5.

\subsection{Preliminary}

The Diffusion Transformer (DiT) model,  uses a transformer as the denoising network to iteratively refine noisy image tokens. A DiT model processes two types of tokens: noisy image tokens $z \in \mathbb{R}^{N \times d}$ and text condition tokens $c_T \in \mathbb{R}^{M \times d}$, where $d$ is the embedding dimension, and $N$ and $M$ are the number of image and text tokens. Throughout the network, these tokens maintain consistent shapes as they pass through multiple transformer blocks.

In FLUX.1, each DiT block consists of layer normalization followed by Multi-Modal Attention (MMA)~\cite{mma}, which incorporates Rotary Position Embedding (RoPE)~\cite{rope} to encode spatial information. For image tokens $z$, RoPE applies rotation matrices based on the token's position $(i,j)$ in the 2D grid:
\begin{equation}
z_{i,j} \rightarrow z=z_{i,j} \cdot R(i,j),
\end{equation}
where $R(i,j)$ is the rotation matrix at position $(i,j)$. Text tokens $c_T$ undergo the same transformation with their positions set to $(0,0)$.

The multi-modal attention mechanism then projects the position-encoded tokens into query $Q$, key $K$, and value $V$ representations. It enables the computation of attention between all tokens:
\begin{equation}
\text{MMA}([z; c_T]) = \text{softmax}\left(\frac{QK^\top}{\sqrt{d}}\right)V,
\end{equation}
where $[z; c_T]$ denotes the concatenation of image and text tokens. This formulation enables bidirectional attention.

\subsection{Overall Architecture}

As shown in Fig. \ref{fig2}, the training of MakeAnything is divided into two stages: First, we train on the MakeAnything dataset using the asymmetric LoRA method, enabling the generation of creative tutorials from text descriptions. Then, the LoRA from this first phase is merged with the Flux base model to form the base model for training the ReCraft Model. In the second stage, image condition tokens are concatenated with noised latent tokens, introducing an image-conditioned mechanism into the denoising process. This setup is further fine-tuned using LoRA to complete the training of the ReCraft Model.

%如图2所示 MakeAnything的训练分为两阶段：首先，我们在MakeAnything数据集上使用不对称LoRA方法进行训练，使其能够从文本描述生成创作教程。接着，第一步中的LoRA与Flux基础模型合并，得到ReCraft模型训练的base model。第二步，通过将image condition tokens 与noised latent tokens 拼接，为去噪过程引入图像条件机制，并通过进一步的LoRA微调实现ReCraft模型的训练。

\subsection{Asymmetric LoRA for Procedural Learning}

% MakeAnything 的核心是将教程的不同帧排列成grid，利用DiT的In-context 能力和注意力机制实现一致性的教程生成。DiT的注意力机制中的令牌倾向于关注空间上相邻的令牌。这种倾向源于扩散模型在预训练过程中捕获的邻近图像像素之间的强相关性。为了提升模型对Grid序列的学习效果，我们提出Serpentine dataset construction方法, 如图2所示，我们将9帧和4帧序列按照蛇形排列成网格，以保证时序上相邻的两帧在空间上也是相邻的（横向相邻或纵向相邻）。

% 另一个挑战是，用所有数据混训一个LORA会导致学习多样化知识时会遇到困难， 仅在单一类型的序列数据上训练LoRA会导致过拟合， 因为每一类过程数据数量有限。为此，我们首次在图像生成中引入不对称的LoRA设计，通过联合训练一个共享的中心矩阵A和多个独立的矩阵B来组合共享知识和专门功能，有效提升了多任务性能。LoRA的每一层由一个A矩阵和一个B矩阵组成，其中A矩阵用于捕捉通用知识，B矩阵则针对特定任务进行适配。这种不对称的LoRA架构可以被表述为：

% 完成训练后，domain-specific矩阵B和domain-agnostic矩阵A组合使用， 兼顾泛化能力和特定任务的性能。本文方法还可以和常规LoRA（非过程LoRA）组合使用， 提升在Unseen domain的效果。

\begin{figure*}[ht]
    \centering
    \includegraphics[width=1.0\linewidth]{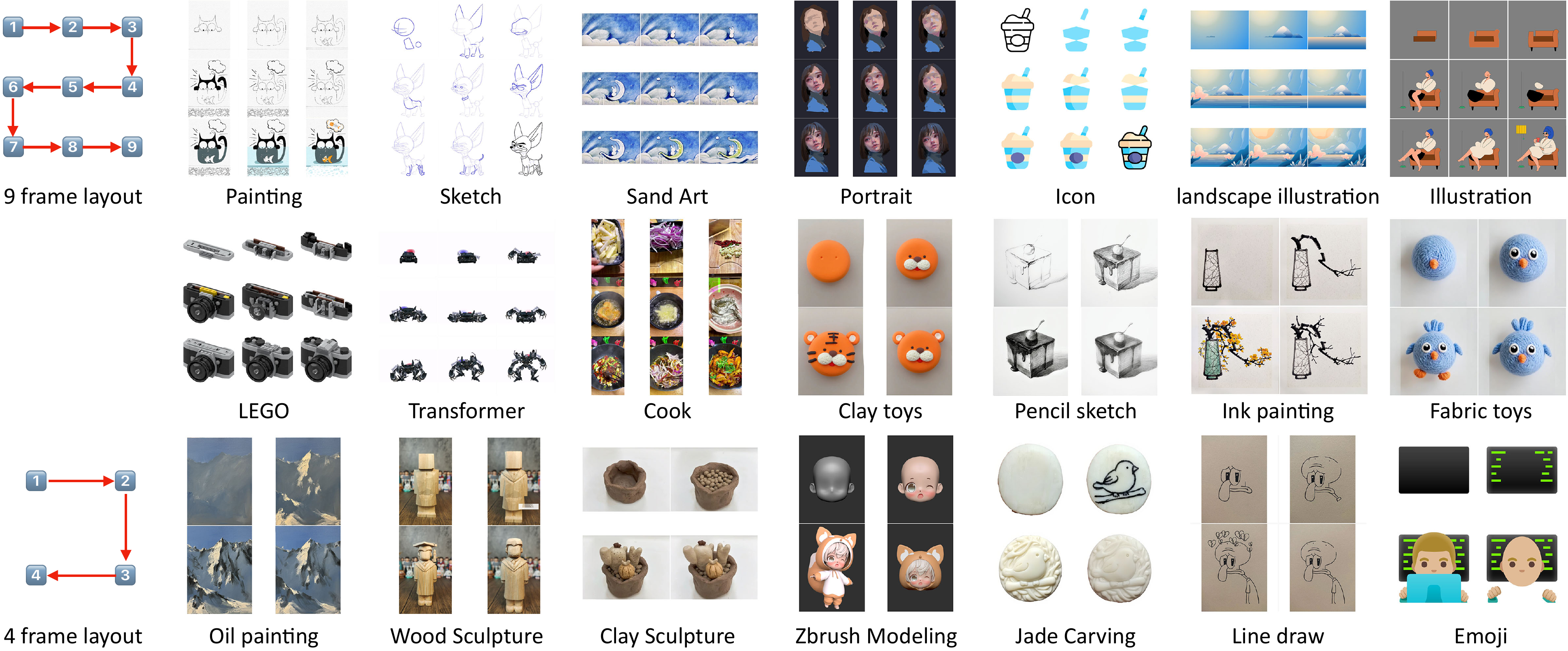} % Replace with your image file
    \caption{Examples from the MakeAnything Dataset, which consists of 21 tasks with over 24,000 procedural sequences.}
    \label{fig3}
\end{figure*}

\noindent \textbf{Serpentine Sequence Layout.}  The core of MakeAnything involves arranging different frames of a sequence into a grid and using the in-context capabilities and attention mechanism of DiT to achieve consistent Sequence generation. Tokens within the DiT's attention mechanism tend to focus on spatially adjacent tokens, a tendency that stems from the strong correlations between neighboring image pixels captured during the pre-training of the diffusion model \cite{grid}. To enhance the model's learning effectiveness for grid sequences, we propose the Serpentine dataset construction method. As shown in Fig. \ref{fig3}, we arrange sequences of 9 frames and 4 frames in a serpentine pattern to ensure that temporally adjacent frames are also spatially adjacent (either horizontally or vertically adjacent).

\noindent \textbf{Asymmetric LoRA.} Another challenge is that training a single LoRA on all data leads to difficulties in learning diverse knowledge, while training LoRA on a single type of sequence data results in overfitting due to the limited quantity of process data for each category. Inspired by HydraLoRA \cite{asymmetry}, we introduce an asymmetric LoRA design for the first time in image generation. This design combines shared knowledge and specialized functionalities by jointly training a shared central matrix \(A\) and multiple task-specific matrices \(B\), significantly improving multi-task performance. 

Each layer of LoRA consists of an \(A\) matrix and a \(B\) matrix, where \(A\) captures general knowledge, and \(B\) adapts to specific tasks. The asymmetric LoRA architecture can be formulated as:
\begin{equation}
W = W_0 + \Delta W = W_0 + \sum_{i=1}^{N} \omega_i \cdot B_iA,
\end{equation}
where \(B_i \in \mathbb{R}^{d \times r}\) and the shared matrix \(A \in \mathbb{R}^{r \times k}\). This structure effectively balances generalization and task-specific adaptation, enhancing the model's performance across diverse tasks.

Inference stage, the domain-specific matrix B and the domain-agnostic matrix A are used in combination, balancing generalization capabilities with performance on specific tasks.  Our method can also be combined with the stylized LoRA from the Civitai website (which is not trained on procedural sequences), to enhance performance in unseen domains.

% 在推理阶段，特定领域的矩阵B和领域无关的矩阵A被组合使用，以平衡泛化能力和特定任务的性能。我们的方法还可以与Civitai网站的风格化LoRA（未在过程序列上训练的）结合使用，以提升在未见领域的性能。

\noindent \textbf{Conditional Flow Matching Loss.} The conditional flow matching loss function is following SD3 \cite{sd3}, which is defined as follows:
\begin{equation}
L_{CFM} = E_{t, p_t(z|\epsilon), p(\epsilon)} \left[ \left\| v_\Theta(z, t, c_T) - u_t(z|\epsilon) \right\|^2 \right]
\end{equation}
Where $ v_\Theta(z, t, c_T)$ represents the velocity field parameterized by the neural network's weights, $t$ is timestep, $c_I$ and $c_T$ are image condition tokens extracted from source image $I_{src}$ and text tokens. $u_t(z|\epsilon)$ is the conditional vector field generated by the model to map the probabilistic path between the noise and true data distributions, and $E$ denotes the expectation, involving integration or summation over time $t$, conditional $z$, and noise $ \epsilon $.

% MakeAnything 数据集的一些例子展示。 MakeAnything包含了21个任务， 共10000+高质量的创建序列。

\subsection{ReCraft Model}
% 在实际应用中，用户除了从文本生成创建过程，更希望能上传一张图片， 预测图片中现有的画作/手工艺品的制作过程。为此，我们实现了ReCraft model， 允许用户上传图像，ReCraft Model生成和上传图像高度一致的过程序列。

% ReCraft model训练的难点是, 对于每一个任务，数据集的数量相当有限，对于从0初始化训练一个可控性插件，如Controlnet或Adapter远远不够。 为此，我们创新的设计了ReCraft model， 通过复用预训练的Flux模型并通过最少的改动，使其扩展为image condition generation model。具体来说，训练环节，我们将最后一帧输入VAE获得latent作为image condition tokens，并与noised latent token 拼起来， 通过注意力机制为其他帧的去噪提供condition信息。 值得一提的是， 加噪和去噪过程仅在其他帧进行，而image condition tokens 是clean的。推理环节，我们通过尾帧预测出前8帧， 即参考图中的物体是如何一步步得到的。

% We aim not only to generate creation processes from text but also to introduce an image-conditioned model capable of predicting the making process of existing artworks or crafts. To achieve this, we developed the ReCraft model, which allows users to upload an image, and the ReCraft model generates a process sequence that closely aligns with the uploaded image.

% The main challenge in training the ReCraft model lies in the limited availability of high-quality sequences. While LoRA is sufficient for small-sample training, initializing a controllable plugin like ControlNet or Adapter from scratch is far from adequate. To address this, we innovatively designed the ReCraft model by repurposing the pre-trained Flux model and adapting it into a conditional generation model that accepts image context. 

% Specifically, during training, we first arrange the creation process sequence into logically ordered 2×2 grid or 3×3 grid. The conditinal image is then fed into a VAE to obtain its latent representation, which is directly appended to the denoising latent at the end.

In practical applications, users not only want to generate creation processes from text but also wish to upload an image and predict the creation process of the existing artwork or handicraft in the picture. For this, we implemented the ReCraft model, which allows users to upload images and generates a sequence of steps highly consistent with the uploaded image.

A major challenge in training the ReCraft model is the limited number of datasets available for each task, which is insufficient to train a controllable plugin like Controlnet or IP-Adapter from scratch. To address this, we innovatively designed the ReCraft model by reusing the pretrained Flux model and making minimal modifications to extend it into an image-conditioned generation model. Specifically, during training, we input the final frame into a VAE to obtain latent image condition tokens, which are then concatenated with noised latent tokens, using the attention mechanism to provide conditional information for denoising other frames. Notably, the noise addition and removal process are only performed on other frames, while the image condition tokens are clean. During inference, we predict the previous eight frames from the tail frame, revealing step-by-step how the object in the reference image was formed.

In ReCraft model, multi-modal attention mechanisms are used to provide conditional information for the denoising of other frames. 
\begin{equation}
\text{MMA}([z; c_I; c_T]) = \text{softmax}\left(\frac{QK^\top}{\sqrt{d}}\right)V,
\end{equation}
where $[z; c_I; c_T]$ denotes the concatenation of image and text tokens. This formulation enables bidirectional attention.

The conditional flow matching loss with image condition can be defined as follows:
\begin{equation}
L_{CFM} = E_{t, p_t(z|\epsilon), p(\epsilon)} \left[ \left\| v_\Theta(z, t, c_I,c_T) - u_t(z|\epsilon) \right\|^2 \right]
\end{equation}
Where $ v_\Theta(z, t, c_I,c_T)$ represents the velocity field parameterized by the neural network's weights.

During inference, Recraft Model predict previous 8 frames based on the final frame. This predicts how the object in the reference image was created step by step.

\begin{figure*}[htp]
    \centering
    \includegraphics[width=1.0\linewidth]{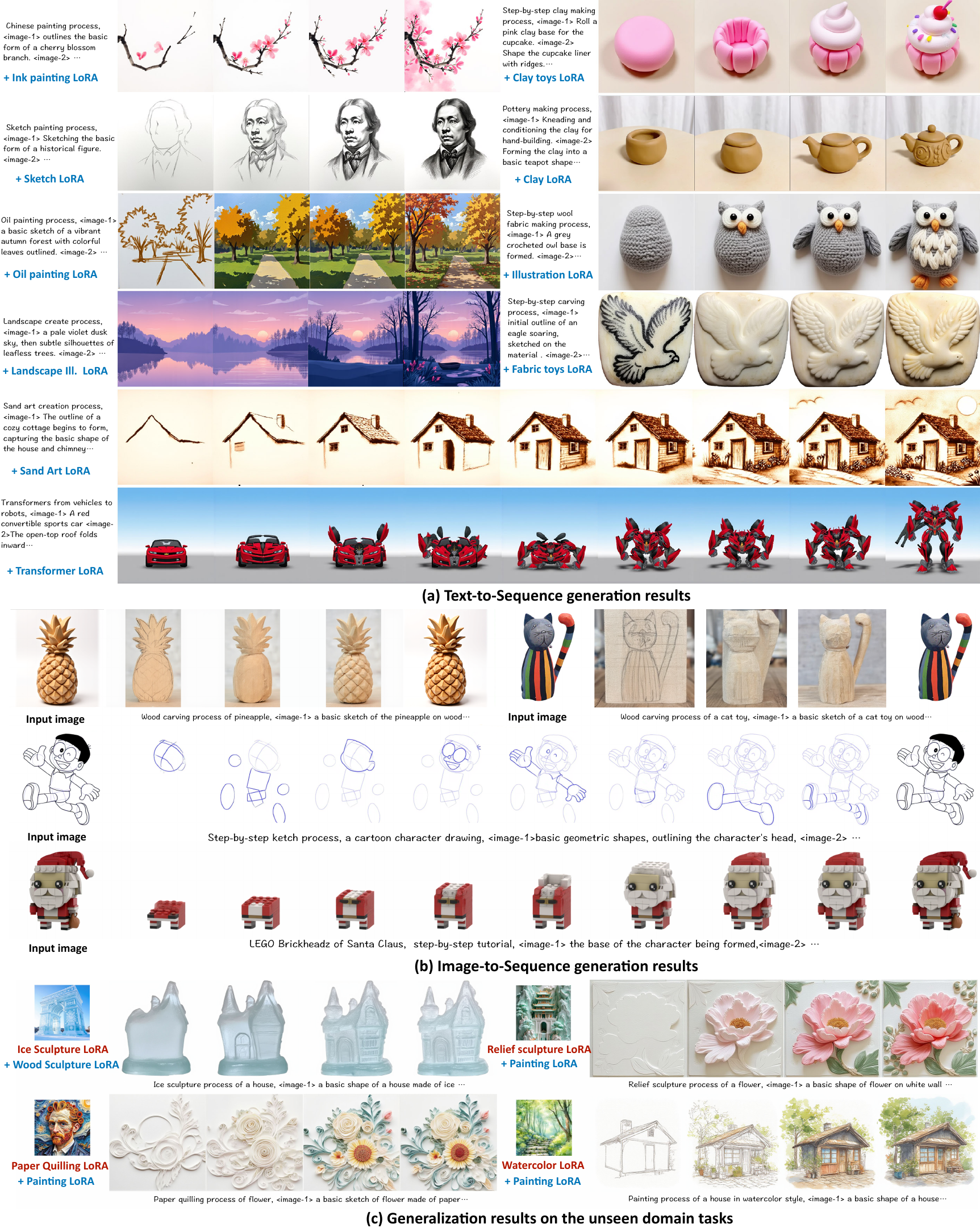} % Replace with your image file
    \vspace{-3mm}
    \caption{Generation results of MakeAnything. From top: \textbf{Text-to-Sequence} outputs conditioned on textual prompts; \textbf{Image-to-Sequence} reconstructions via ReCraft Model; \textbf{Unseen Domain} generalization combining procedural LoRA (blue) with stylistic LoRA (red).  }
    \label{fig4}
\end{figure*}
\vspace{-1mm}

\begin{figure*}[htp]
    \centering
    \includegraphics[width=0.99\linewidth]{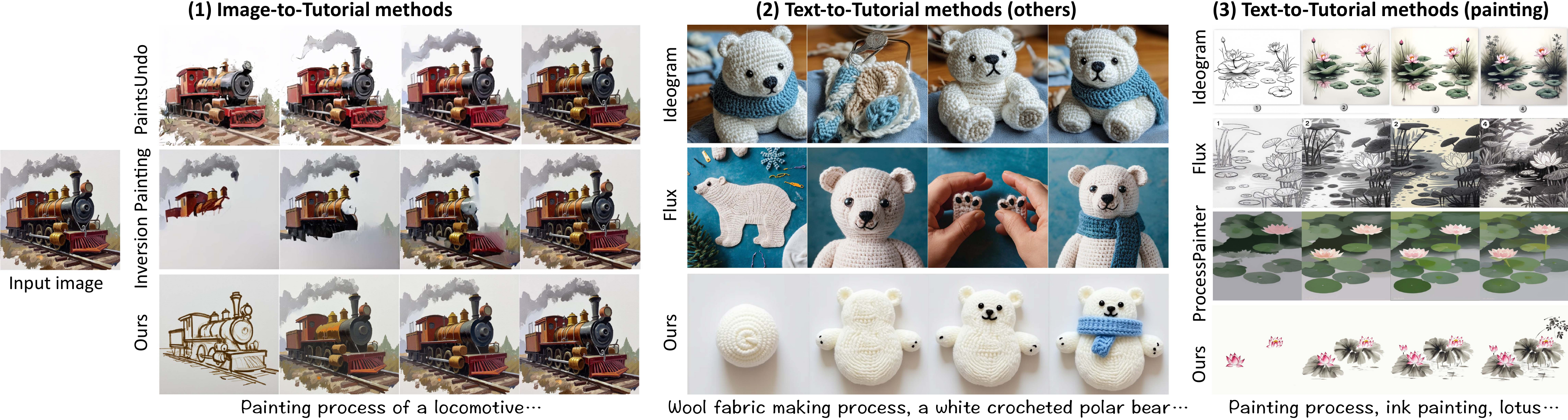} % Replace with your image file
    \vspace{-3mm}
    \caption{Compare with baselines on different tasks.}
    \label{fig5}
\end{figure*}

\subsection{MakeAnything Dataset}
 
% 如图3所示，我们收集了一个多任务教程数据集， 包含21个任务的教程， 类别分别是Painting,  Sketch, Sand Art, Portrait, Icon, Landscape illutsration, illutsration, LEGO, Transformer, Cook, Clay toys, Pencil sketch, Chinese painting, Fabric toys, Oil painting, Wood Sculpture, Clay Sculpture, Brush Modeling, Jade Carving, Line draw, Emoji。我们组建了专业的数据收集和标注小组，从互联网上收集并后处理各种教程， 我们还和艺术家合作，定制高质量绘画过程数据。 不同任务的数据量不同，50至10000不等，共计24,000+. 前十个任务数据是9帧， 其他的是4帧， 分别被组成3*3和2*2的grid用于训练。我们通过GPT4-o为所有数据集打文本标签， 并对每一帧进行描述。 

As shown in Fig. \ref{fig3},  we have collected a multi-task dataset that encompasses tutorials for 21 tasks. We assembled a professional data collection and annotation team that gathered and processed various tutorials from the internet and also collaborated with artists to customize high-quality painting process data. The datasets vary in size from 50 to 10,000 entries, totaling over 24,000. The first ten tasks have data in 9 frames, while the rest have 4 frames, arranged into 3x3 and 2x2 grids for training purposes. We used GPT-4o to label all datasets and describe each frame.

\section{Experiment}

\subsection{Experimental Setting}
% 我们基于预训练的Flux 1.0 dev 实现了MakeAnything. 我们基于Kohayss的Flux LoRA训练代码，并将优化器从Adam优化器换为CAME优化器，实验发现这一设置能获得更好的生成质量。不对称LoRA和ReCraft model训练阶段，分辨率被设置为1024*1024， LoRA rank 64， 学习率 1e-4， batch size 2。 不对称LoRA和ReCraft model分别训练40000step和15000step。 

\textbf{Setup.} We implemented MakeAnything based on the pre-trained Flux 1.0 dev. We replaced the Adam optimizer with the CAME optimizer, and experiments showed that this setup achieved better generation quality. During the training phases of Asymmetric LoRA and the ReCraft model, the resolution was set to 1024, LoRA rank was 64, learning rate was 1e-4, and batch size was 2. Asymmetric LoRA and the ReCraft model were trained for 40,000 steps and 15,000 steps, respectively.

% The Strategy Adapter leverages the pre-trained model of InstantX Flux IP-Adapter. The number of image tokens is set to 128, and the model is trained on the 100k dataset with a batch size of 128 over 30k training steps.

\textbf{Baselines.} In the Text-to-Sequence task, we compare our approach with state-of-the-art baseline methods, namely ProcessPainter \cite{processpainter}, Flux 1.0 \cite{flux}, and the commercial API Ideogram \cite{ideogram}. We categorize the test prompts into two types: painting and others, because some baselines only support painting. In the Image-to-Sequence task, our baselines are Inverse Painting \cite{inverse} and PaintsUndo \cite{paintsundo}, which are capable of predicting the creation process of a painting.

% 在Text-to-Tutorial 任务中， 我们和最先进的baseline方法对比，分别是ProcessPainter，Flux 1.0, 商业API Ideogram，我们将测试prompt分为绘画和其他两类，因为部分baseline仅支持绘画。
% 在Image-to-Tutorial 任务中，我们对比的baseline是Inverse Painting 和 PaintsUndo， 可以预测画作的创作过程。

\textbf{Evaluation Metrics.} A good procedural sequence needs to be coherent, logical, and useful; however, evaluating procedural sequence generation and its rationality lacks precedents. We employ the CLIP Score to assess the text-image alignment of the generated results. Additionally, we evaluate the coherence and usability of the generated results using GPT-4o and human evaluations. Specifically, we meticulously design the input prompts for GPT-4o and scoring rules to align with human preferences.  In the comparative experiments, we concatenate the results from various baselines with our results, input them into GPT-4o all at once, and have it select the best results across different evaluation dimensions.

\begin{table}[ht]
\centering
\tiny % Makes the font smaller than \footnotesize
\caption{Combined Evaluation of Procedural Sequence Generation Results Across Different Tasks. Abbreviations: G = GPT score, H = Human evaluation, C = CLIP score.}
\label{tab:task_evaluation}
\begin{tabular}{l c c c}
\toprule % Top line
\textbf{Task} & \textbf{Alignment (G \textbar\ H \textbar\ C)} & \textbf{Coherence (G \textbar\ H)} & \textbf{Usability (G \textbar\ H)} \\ 
\midrule
Painting & 4.50 \textbar\ 4.27 \textbar\ 34.24 & 4.80 \textbar\ 3.98 & 4.60 \textbar\ 4.13 \\
Sketch & 4.10 \textbar\ 3.97 \textbar\ 29.35 & 4.70 \textbar\ 4.11 & 4.10 \textbar\ 4.13 \\
Sand Art & 4.20 \textbar\ 4.30 \textbar\ 31.82 & 4.70 \textbar\ 4.12 & 4.30 \textbar\ 4.18 \\
Portrait & 4.25 \textbar\ 4.28 \textbar\ 33.84 & 5.00 \textbar\ 4.28 & 4.05 \textbar\ 4.33 \\
Icon & 3.45 \textbar\ 4.33 \textbar\ 31.46 & 3.50 \textbar\ 4.17 & 3.15 \textbar\ 4.25 \\
Landscape Ill. & 4.55 \textbar\ 4.28 \textbar\ 32.25 & 4.85 \textbar\ 3.95 & 4.50 \textbar\ 4.12 \\
Illustration & 3.12 \textbar\ 4.17 \textbar\ 31.68 & 3.40 \textbar\ 4.07 & 2.45 \textbar\ 4.07 \\
LEGO & 4.60 \textbar\ 4.32 \textbar\ 34.40 & 4.90 \textbar\ 4.15 & 4.75 \textbar\ 4.00 \\
Transformer & 4.75 \textbar\ 4.30 \textbar\ 33.03 & 4.90 \textbar\ 4.23 & 4.75 \textbar\ 4.15 \\
Cook & 3.20 \textbar\ 4.21 \textbar\ 34.41 & 4.25 \textbar\ 4.03 & 3.65 \textbar\ 3.90 \\
Clay Toys & 4.30 \textbar\ 4.17 \textbar\ 35.25 & 4.50 \textbar\ 4.30 & 4.20 \textbar\ 4.30 \\
Pencil Sketch & 3.85 \textbar\ 4.33 \textbar\ 34.44 & 4.50 \textbar\ 4.20 & 3.80 \textbar\ 4.25 \\
Chinese Painting & 4.80 \textbar\ 4.37 \textbar\ 33.46 & 4.90 \textbar\ 4.22 & 4.70 \textbar\ 4.33 \\
Fabric Toys & 4.35 \textbar\ 4.30 \textbar\ 32.83 & 4.60 \textbar\ 4.08 & 4.40 \textbar\ 4.30 \\
Oil Painting & 4.90 \textbar\ 4.30 \textbar\ 37.30 & 4.95 \textbar\ 4.17 & 4.85 \textbar\ 4.20 \\
Wood Sculpture & 4.65 \textbar\ 4.32 \textbar\ 33.83 & 4.85 \textbar\ 4.23 & 4.65 \textbar\ 4.08 \\
Clay Sculpture & 4.30 \textbar\ 4.17 \textbar\ 35.25 & 4.50 \textbar\ 4.30 & 4.20 \textbar\ 4.30 \\
Brush Modeling & 4.20 \textbar\ 4.33 \textbar\ 32.27 & 4.15 \textbar\ 4.03 & 4.05 \textbar\ 4.25 \\
Jade Carving & 4.90 \textbar\ 4.28 \textbar\ 32.93 & 4.85 \textbar\ 4.12 & 4.75 \textbar\ 4.00 \\
Line Draw & 4.10 \textbar\ 4.20 \textbar\ 30.76 & 4.20 \textbar\ 3.97 & 3.90 \textbar\ 4.08 \\
Emoji & 3.75 \textbar\ 4.25 \textbar\ 34.20 & 3.60 \textbar\ 4.17 & 3.80 \textbar\ 4.18 \\
\bottomrule % Bottom line
\label{tab1}
\end{tabular}
\end{table}

\subsection{Experimental Results}
% 图 \ref{fig4}(a) 展示了从文本描述生成过程序列的结果。得益于高质量数据集、健壮的预训练模型和创新的方法设计，MakeAnything一致地产生了高质量和逻辑连贯的过程序列。表1展示了MakeAnything在21个任务上的定量评估结果，包括GPT和人工评分， 每个ren w

% 图 \ref{fig4}(b) 突出显示了模型基于输入图像生成过程序列的能力。结果显示生成的序列与原始图像内容高度一致。这展示了模型解释复杂视觉输入并重构逻辑一致的创作过程的能力，使其能够在逆向工程和教育教程等多个领域中应用。

Fig. \ref{fig4}(a) showcases the results of generating process sequences from textual descriptions. Benefiting from high-quality datasets, a robust pre-trained model, and an innovative method design, MakeAnything consistently produces high-quality and logically coherent process sequences. Table \ref{tab1} presents the quantitative evaluation results of MakeAnything across 21 tasks, including scores from GPT and human assessments, with 20 sequences generated per task, with 20 sequences generated per task.

Fig. \ref{fig4}(b) highlights the model's ability to generate process sequences conditioned on input images. The results indicate a high degree of alignment between the generated sequences and the original image content. This showcases the model's capacity to interpret complex visual inputs and reconstruct logically consistent creation processes, enabling its application in diverse fields such as reverse engineering and educational tutorials.

% 图4 (c) 展示了MakeAnything 在unseen domain 上的结果，我们从civitai网站收集了 水彩、浮雕、冰雕、衍纸画等LoRA，并和我们的教程LoRA组合使用。可以看出，MakeAnything展示出相当不错的泛化能力，尽管训练时没有见过这些题材的创作过程。

Fig. \ref{fig4}(c) shows the results of MakeAnything in unseen domains. We collected various LoRAs from the Civitai \cite{civitai2025} website, including watercolor, relief, ice sculpture, and paper quilling art, and combined them with our procedural LoRA. It is evident that MakeAnything demonstrates quite impressive generalization capabilities, despite not having been trained on these creative processes.

\begin{figure*}[htp]
    \centering
    \includegraphics[width=1.0\linewidth]{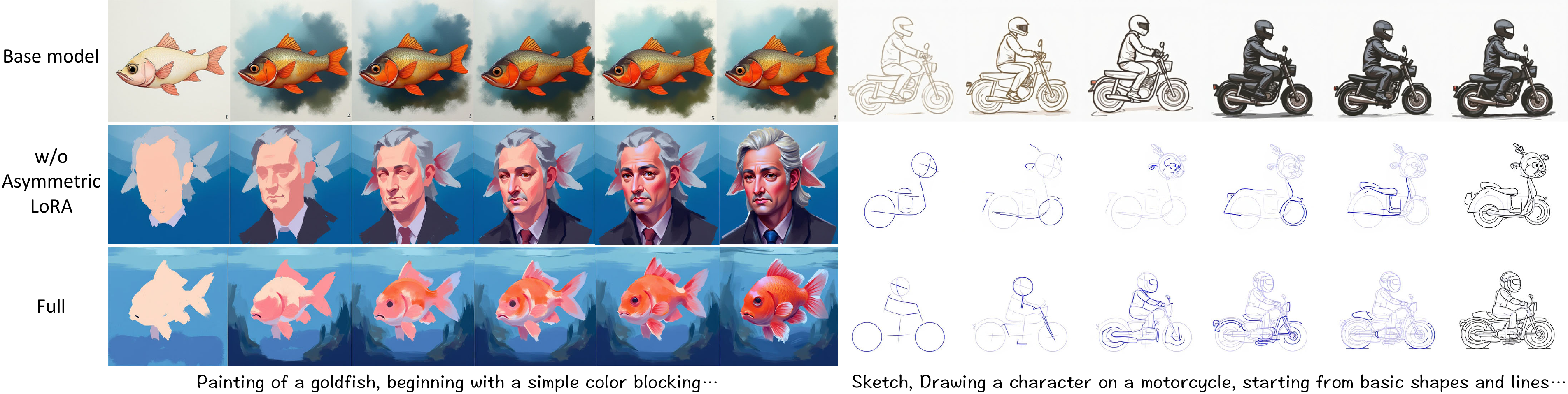} % Replace with your image file
    \vspace{-4mm}
    \caption{Ablation study results.}
    \vspace{-4mm}
    \label{fig6}
\end{figure*}

\begin{figure}[htp]
    \centering
    \includegraphics[width=1.0\linewidth]{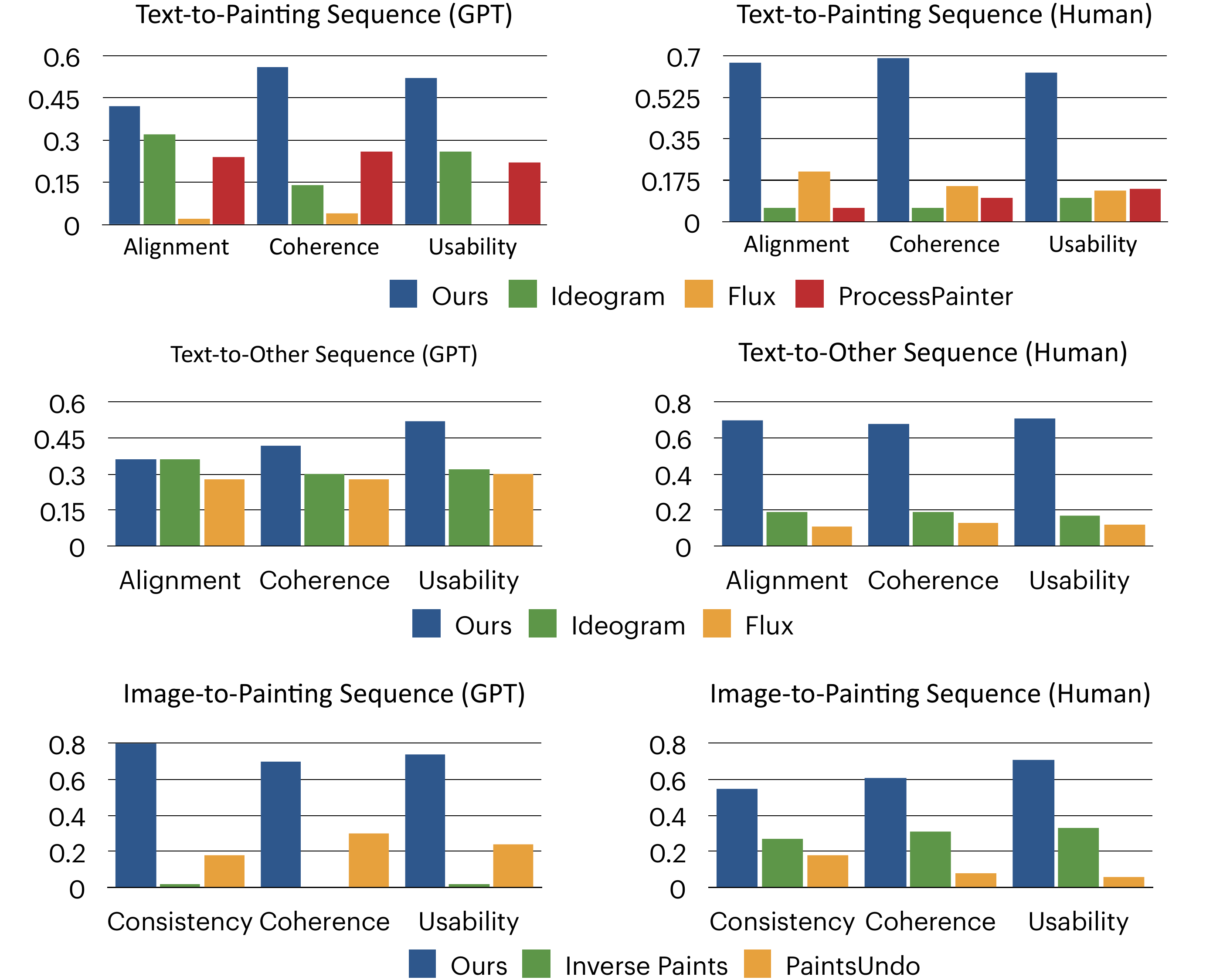} % Replace with your image file
    \caption{Comparison results on three tasks, evaluated by GPT and humans respectively.}
    \vspace{-2mm}
    \label{fig7}
\end{figure}

\subsection{Comparation and  Evaluation} 
This section consolidates the comparative evaluations of our method against baseline approaches on 50 sequence groups. Fig. \ref{fig5}(a) and (b), demonstrate that MakeAnything produces higher quality procedural sequence with superior logic and coherence, unlike the baseline methods which struggle with consistency. Fig. \ref{fig5}(c) compares the ReCraft model to a baseline, highlighting our method's training on diverse real data, resulting in varied and authentic creative processes. Quantitative results in  Fig. \ref{fig7} confirm MakeAnything's superiority in Text-Image Alignment, Coherence, and Usability across all tested metrics.

% 这一节整合了我们的方法与基线方法的比较评估， 测试集的为50组序列. 如图5(a)和(b)所示，MakeAnything生成的教程质量更高，逻辑和连贯性也更好，而基线方法则在一致性上存在问题。图5(c)展示了ReCraft模型与基线方法的比较结果，我们的方法在多样的真实数据上进行训练，而不仅仅是合成数据或人类绘画序列的模拟，这使得生成的绘画过程更加多样化且贴近艺术家的实际创作过程。图6的定量评估结果也证明了MakeAnything在文本-图像对齐、连贯性和可用性所有测试指标上的领先。

% This section presents the comparative results between our method and the baseline approaches. As shown in Fig. \ref{fig5}(a) and (b), MakeAnything generates tutorials of higher quality, whereas the baseline methods fail to produce consistent procedural sequences, lacking in logic and coherence.  Fig. \ref{fig6} displays the quantitative evaluation results, where MakeAnything excels in Text-Image Alignment, Coherence, and Usability.

% Fig. \ref{fig5}(c) illustrates the comparison results between the ReCraft model and the baseline method. The baseline method employs a uniform process strategy, whereas our method is trained on a diverse set of real data, not synthetic data or simulations of human painting sequences. This enables the generation of various types of painting processes that align more closely with artists' actual creative processes. Fig. \ref{fig6} shows the quantitative evaluation results, with our method leading across all metrics.

% 本节展示本文方法和baseline方法的对比结果， 如图5所示， MakeAnything 教程生成结果质量更高，baseline方法并不能稳定的生成过程序列，逻辑性和一致性不足。表3展示了定量评估结果，MakeAnything在 Text-Image Align， Coherence， 和Usability上取得了最好的结果。 

% 图5展示了ReCraft model和baseline method的对比结果，baseline 方法生成的过程策略单一， 我们的方法在多样的真实数据上训练，而不是合成数据或对人类绘画顺序的模拟，能够生成不同种类的绘画过程， 结果和画家的创作过程更一致。 图6 展示了GPT和人类评估结果， 我们的方法在所有指标上取得领先。

\subsection{User Study}
% 为了充分研究MakeAnything的有效性和改进方向， 我们进行了详细的用户研究。我们设计了问卷，将本文方法的结果和baseline方法的结果同时展示，让用户选择 Alignment、Coherence、Usability、Consistency 更好的序列。图7展示了用户研究结果，MakeAnything的结果全面领先。此外我们收集了用户对MakeAnything的改进建议。一些用户表示，创建过程希望是图文并茂的，加入必要的文字介绍。一些用户认为更详细的过程是有用的，4帧的创建过程在有些任务上不够详细。

% To thoroughly evaluate the effectiveness of {MakeAnything and identify potential areas for improvement, we conducted a detailed user study. A questionnaire was designed to gather user preferences, asking participants to select the most preferred and the most helpful process sequences. Fig. \ref{fig7} presents the results of the user study, showing that MakeAnything outperformed all competitors comprehensively. 

To comprehensively evaluate MakeAnything's effectiveness, we conducted a user study comparing our method against baselines. Participants rated sequences across four metrics: Alignment (text-image similarity), Coherence (logical step progression), Usability (practical value), and Consistency (consistency between image condition). As shown in Fig. \ref{fig7}, MakeAnything demonstrates comprehensive superiority across all metrics.

%此外我们收集了用户对MakeAnything的改进建议。一些用户表示，创建过程希望是图文并茂的，加入必要的文字介绍。一些用户认为更详细的过程是有用的，4帧的创建过程在有些任务上不够详细。

% Additionally, we collected user feedback on how to improve MakeAnything. Some users have expressed a desire for the creation process to be illustrated with both images and necessary textual descriptions. Some users believe that a more detailed process is useful, as a 4-frame creation process is not sufficiently detailed for some tasks.

\subsection{Ablation Study} 
% 本节，我们对 不对称LoRA的进行了消融实验，图6对比了肖像生成 和 sketch生成的效果。前者在50张肖像绘画序列上训练， 后者在300张卡通角色 sketch 序列上训练。我们对比了，base model 的结果、标准LoRA的结果，和采用对不对称LoRA，共享A矩阵的结果。从结果可以看出，尽管base model无法生成合理的分步骤结果，但是text following整体不错。采用标准LoRA在类别分布不均匀的小数据上训练导致了严重的过拟合，虽然分步骤的过程合理，text-image alignment 显著变差。 而采用不对称LoRA结果很好的兼顾过程合理性和text-image alignment。我们认为在海量过程数据上训练的A矩阵学习到了更多通用的知识，有利于缓解过拟合。表2展示了在更多任务上的定量实验结果，进一步证实结论。

% In this section, we conducted ablation experiments on asymmetric LoRA, and Fig. \ref{fig6} compares the results of portrait and sketch tutorial generation task. The former was trained on 50 portrait painting sequences, while the latter was trained on 300 cartoon character sketch sequences. We compared the results of the base model, standard LoRA, and asymmetric LoRA with a shared A matrix. The results show that although the base model fails to generate reasonable step-by-step results, the text following is overall quite good. Training standard LoRA on small datasets with uneven category distribution led to severe overfitting; although the step-by-step process was reasonable, text-image alignment significantly worsened. In contrast, using asymmetric LoRA effectively balanced process rationality and text-image alignment. We believe that the A matrix, trained on a massive amount of procedural data, learned more general knowledge, helping to mitigate overfitting. Table. \ref{ab} presents quantitative experimental results on additional tasks, further confirming the conclusions.

In this section, we conducted ablation experiments on asymmetric LoRA, and Fig. \ref{fig6} compares the results of portrait and sketch tutorial generation task. The former was trained on 50 portrait painting sequences, while the latter was trained on 300 cartoon character sketch sequences. While the base model produces coherent text but fails in step-by-step synthesis, standard LoRA exhibits severe overfitting on small datasets with imbalanced class distributions—yielding plausible steps but compromised text-image alignment. Our method achieves both procedural rationality and text-image alignment by leveraging knowledge from large-scale pretraining. Quantitative results across more tasks (Table~\ref{ab}) further validate these findings.

\begin{table}[ht]
\centering
\tiny % Makes the font smaller than \footnotesize
\caption{Ablation Study Results Using GPT Evaluation and CLIP Score.}  
\label{tab:task_evaluation}
\begin{tabular}{ccccc}
\toprule % Top line
\textbf{Model} & \textbf{Task} & \textbf{Alignment(G \textbar\ C)} & \textbf{Coherence} & \textbf{Usability} \\ 
\midrule % Middle line
\multirow{3}{*}{Base Model} & Portrait & 3.75\textbar\ 29.78 & 3.45 & 3.35 \\
                                & Wood & 3.25\textbar\ \textbf{35.29}  & 2.95 & 2.65 \\
                                & Fabric toys & 3.55\textbar\ \textbf{32.95}  & 4.00 & 3.85 \\
\midrule % Middle line
\multirow{3}{*}{w/o Asymmetric LoRA} & Portrait & 4.25\textbar\ 31.08  & 4.50 & 4.15 \\
                                & Wood Sculpture & 3.55\textbar\ 31.05  & \textbf{4.35} & 3.75 \\
                                & Fabric toys & 3.75\textbar\ 30.72  & 3.15 & 3.20 \\
\midrule % Separate section
\multirow{3}{*}{Full} & Portrait & \textbf{4.55}\textbar\ \textbf{32.95}  & \textbf{4.75} & \textbf{4.25} \\
                      & Wood Sculpture & \textbf{4.25}\textbar\ 33.89  & 3.80 & \textbf{4.05} \\
                      & Fabric toys & \textbf{4.40}\textbar\ 32.01  & \textbf{4.25} & \textbf{4.35} \\
\bottomrule % Bottom line
\label{ab}
\end{tabular}
\end{table}

\section{Limitations and Future Work}
The current grid-based composition strategy in MakeAnything introduces two inherent limitations: constrained output resolution (max 1024×1024) and fixed frame count (up to 9 steps). We plan to address these limitations in future work, enabling arbitrary-length sequence generation with high-fidelity outputs.

% MakeAnything中当前的基于网格的构图策略引入了两个固有的限制：受限的输出分辨率（最高1024×1024）和固定的帧数（最多9步）。这些限制源于我们的设计选择，通过空间平铺来统一多帧生成。我们计划在未来的工作中解决这些限制，从而实现具有高保真输出的任意长度序列生成。

\section{Conclusion}
We introduced MakeAnything, a novel framework for generating high-quality process sequences using the DiT model with LoRA fine-tuning. By leveraging multi-domain procedural dataset and adopting an asymmetric LoRA design, our approach effectively balances generalization and task-specific performance. Additionally, the image-conditioned plugin enables controllable and interpretable sequence generation. Extensive experiments demonstrated the superiority of our method across diverse tasks, establishing a new benchmark in this field. Our contributions pave the way for further exploration of step-by-step process generation, opening up exciting possibilities in computer vision and related applications.

% 在本文中，我们提出了MakeAnything，这是一种通过DiT模型结合LoRA微调生成高质量过程序列的新颖框架。通过利用精心收集的多领域数据集并采用不对称LoRA设计，我们的方法有效地平衡了泛化能力和任务特定性能。此外，图像条件插件支持可控且可解释的序列生成。广泛的实验表明，我们的方法在各种任务上表现出色，确立了该领域的新基准。我们的贡献为进一步探索逐步过程生成开辟了新途径，为计算机视觉及相关应用带来了激动人心的可能性。

% 我们提出了MakeAnything， 可以从

\nocite{langley00}

\newpage
\section*{Impact Statement}
This paper presents work whose goal is to advance the field of Machine Learning. There are many potential societal consequences of our work, none which we feel must be specifically highlighted here.

\bibliography{icml2025/main}
\bibliographystyle{icml2025}

%%%%%%%%%%%%%%%%%%%%%%%%%%%%%%%%%%%%%%%%%%%%%%%%%%%%%%%%%%%%%%%%%%%%%%%%%%%%%%%
%%%%%%%%%%%%%%%%%%%%%%%%%%%%%%%%%%%%%%%%%%%%%%%%%%%%%%%%%%%%%%%%%%%%%%%%%%%%%%%
% APPENDIX
%%%%%%%%%%%%%%%%%%%%%%%%%%%%%%%%%%%%%%%%%%%%%%%%%%%%%%%%%%%%%%%%%%%%%%%%%%%%%%%
%%%%%%%%%%%%%%%%%%%%%%%%%%%%%%%%%%%%%%%%%%%%%%%%%%%%%%%%%%%%%%%%%%%%%%%%%%%%%%%
\newpage
\appendix
\onecolumn

\section{Implementation details of the GPT4-o evaluation.}

In the GPT-4-o evaluation process, we tailor distinct evaluation metrics for different tasks, ensuring both direct scoring and selective ranking are covered to suit the task's nature.

\subsection{Direct Scoring Evaluation (for Procedural Sequence Generation and Ablation Studies)}
The assistant evaluates a sequence of images depicting a procedural process with criteria such as:
\begin{itemize}
  \item \textbf{Accuracy:} Measures content alignment with the provided prompt, scored from 1 (not accurate) to 5 (completely accurate).
  \item \textbf{Coherence:} Assesses logical flow from 1 (disjointed) to 5 (seamless progression).
  \item \textbf{Usability:} Rates helpfulness for understanding the procedure from 1 (not helpful) to 5 (highly helpful).
\end{itemize}
Scores are output in JSON format, for example:
\begin{verbatim}
{
  "Accuracy": 4,
  "Coherence": 5,
  "Usability": 4
}
\end{verbatim}

\subsection{Selective Ranking Evaluation (for User Study Comparisons)}
This evaluation compares multiple images from different models, ranking them by:
\begin{itemize}
  \item \textbf{Accuracy:} Which image best represents the prompt?
  \item \textbf{Coherence:} Which image shows the clearest, most logical process?
  \item \textbf{Usability:} Which image offers the most helpful visual guidance?
\end{itemize}
Rankings are provided from 1 (best) to 4 and outputted in JSON format, e.g.,
\begin{verbatim}
{
  "Accuracy": 1,
  "Coherence": 2,
  "Usability": 3
}
\end{verbatim}

\textbf{Example of Task Prompt and Evaluation:}
Prompt: "This image shows the process of creating a handmade sculpture."
Images: [Upload images of models 1, 2, 3, and 4]
Evaluation: The assistant ranks the models for Accuracy, Coherence, and Usability in JSON format.
This evaluation merges qualitative and quantitative assessments to determine the effectiveness of the images generated by GPT-4-o models.

\section{More results}

% 表3-表6 展示GPT评估和人类评估的原始数据。
Fig 8-11 show more generation results of MakeAnything. Table 3-6 display the raw data from GPT evaluations and human assessments.

\begin{table}[htp]
\centering % Makes the font smaller than \footnotesize
\scriptsize
\caption{Compare with Text-to-Sequence methods (GPT)}
\label{tab:task_evaluation}
\begin{tabular}{c c c c c c}
\toprule % Top line
\textbf{Category} & \textbf{Methods} & \textbf{Alignment} & \textbf{Coherence} & \textbf{Usability} \\ 
\midrule % Middle line
\multirow{4}{*}{Painting} & Processpainter & 0.24 & 0.26 & 0.22 \\
                                & Ideogram & 0.32 & 0.14 & 0.26 \\
                                & Flux & 0.02 & 0.04 & 0.00 \\
                                & Ours & \textbf{0.42} & \textbf{0.56} & \textbf{0.52} \\
\midrule % Separate section
\multirow{3}{*}{Others} & Ideogram & 0.36 & 0.30 & 0.32 \\
                      & Flux & 0.28 & 0.28 & 0.30 \\
                      & Ours & \textbf{0.36} & \textbf{0.42} & \textbf{0.38} \\
\bottomrule % Bottom line
\label{tab3}
\end{tabular}
\end{table}

\begin{table}[htp]
\centering% Makes the font smaller than \footnotesize
\scriptsize
\caption{Compare with Image-to-Sequence methods (GPT)}
\begin{tabular}{c c c c c c}
\toprule % Top line
\textbf{Category} & \textbf{Methods} & \textbf{Consistency} & \textbf{Coherence} & \textbf{Usability} \\ 
\midrule % Middle line
\multirow{3}{*}{Painting} & Inverse Paints & 0.02 & 0.00 & 0.02 \\
                                &PaintsUndo  & 0.18 & 0.30 & 0.24 \\
                                &Ours & \textbf{0.80} & \textbf{0.70} & \textbf{0.74} \\
\bottomrule % Bottom line
\label{tab4}
\end{tabular}
\end{table}

\begin{table}[htb]
\centering% Makes the font smaller than \footnotesize
\scriptsize
\caption{Compare with Text-to-Sequence methods (Human)}
\begin{tabular}{c c c c c c}
\toprule % Top line
\textbf{Category} & \textbf{Methods} & \textbf{Alignment} & \textbf{Coherence} & \textbf{Usability} \\ 
\midrule % Middle line
\multirow{4}{*}{Painting} & Processpainter & 0.06 & 0.10 & 0.14   \\
                                & Ideogram & 0.06 & 0.06  & 0.10 \\
                                & Flux & 0.21 & 0.15 & 0.13 \\
                                & Ours & \textbf{0.67} & \textbf{0.69} & \textbf{0.63} \\
\midrule % Separate section
\multirow{3}{*}{Others} & Ideogram & 0.19 & 0.19 & 0.17 \\
                      & Flux & 0.11 & 0.13 & 0.12 \\
                      & Ours & \textbf{0.70} & \textbf{0.68} & \textbf{0.71} \\
\bottomrule % Bottom line
\label{tab5}
\end{tabular}
\end{table}

\begin{table}[h!]
\centering% Makes the font smaller than \footnotesize
\scriptsize
\caption{Compare with Image-to-Sequence methods (Human)}
\begin{tabular}{c c c c c c}
\toprule % Top line
\textbf{Category} & \textbf{Methods} & \textbf{Consistency} & \textbf{Coherence} & \textbf{Usability} \\ 
\midrule % Middle line
\multirow{3}{*}{Painting} & Inverse Paints & 0.27 & 0.31  & 0.33 \\
                                &PaintsUndo  & 0.18 & 0.08 & 0.06 \\
                                &Ours & \textbf{0.55} & \textbf{0.61} & \textbf{0.61}  \\
\bottomrule % Bottom line
\label{tab6}
\end{tabular}
\end{table}

\begin{figure*}[htp]
    \centering
    \includegraphics[width=0.85\linewidth]{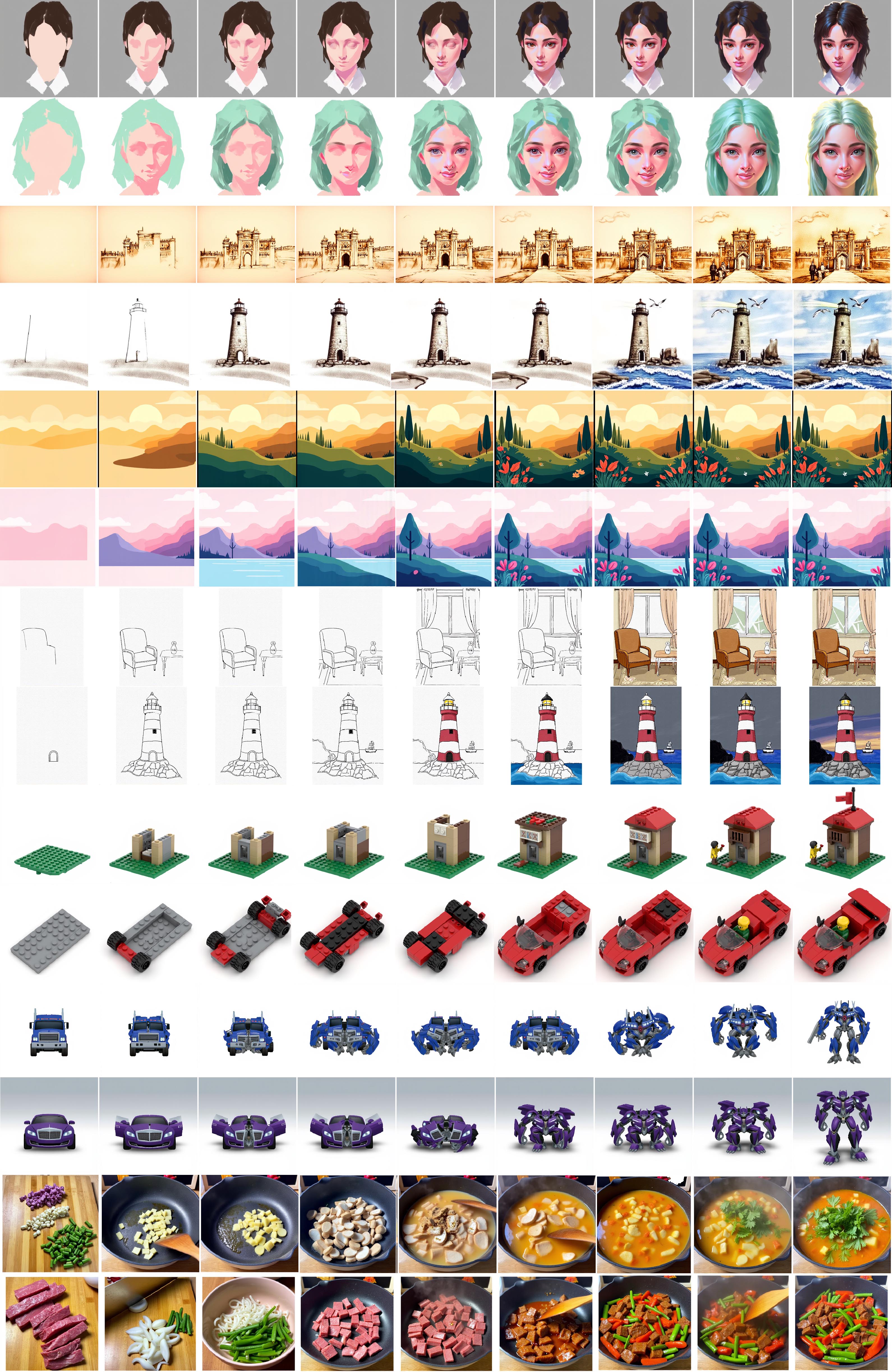} % Replace with your image file
    \caption{More generation results. From top to bottom, they are portrait, Sand Art, landscape illustration, painting, LEGO, transformer, and cook respectively.}
    \label{fig8}
\end{figure*}

\begin{figure*}[htp]
    \centering
    \includegraphics[width=0.85\linewidth]{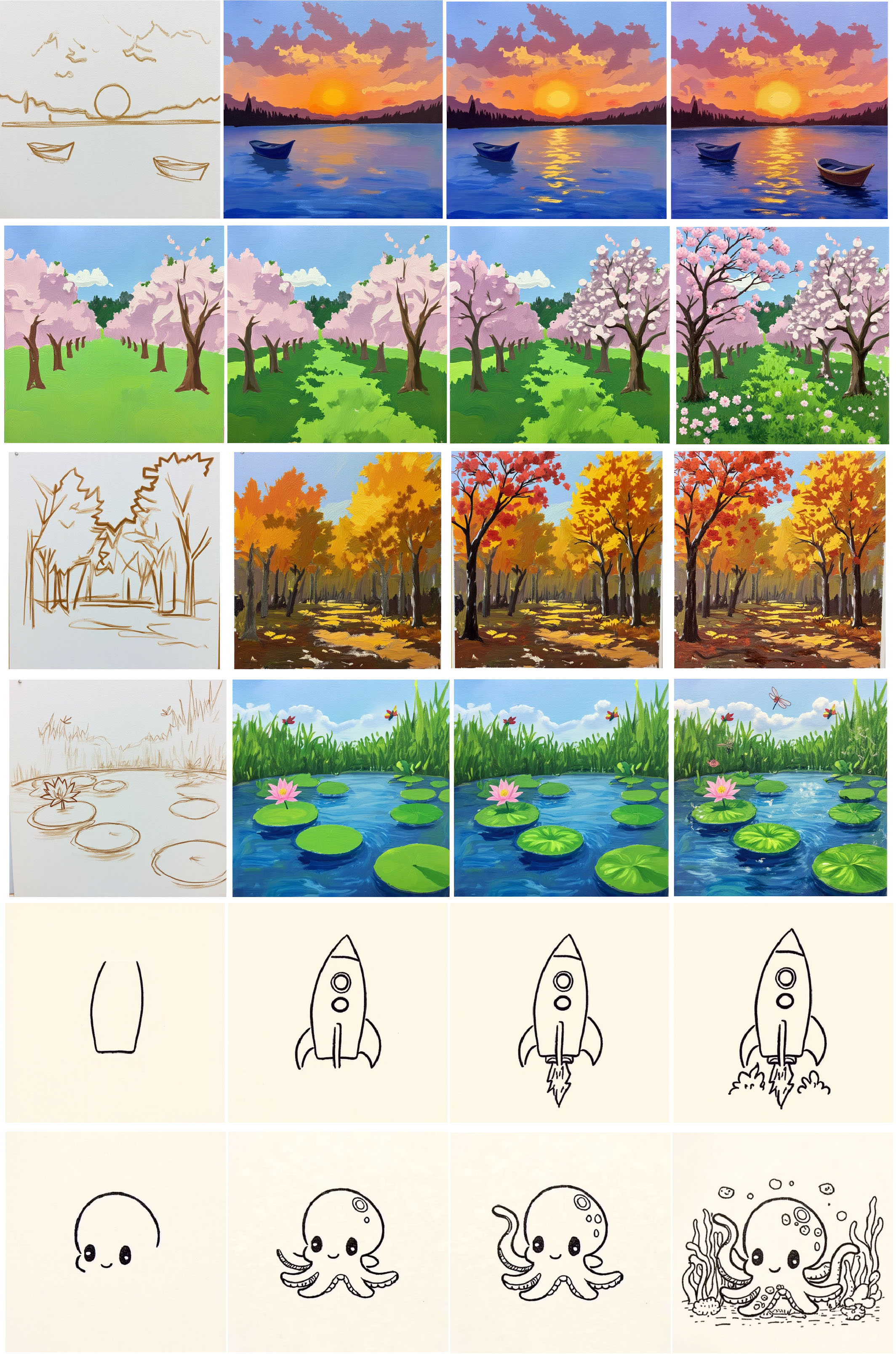} % Replace with your image file
    \caption{More generation results. From top to bottom, they are oil painting and line draw.}
    \label{fig9}
\end{figure*}

\begin{figure*}[htp]
    \centering
    \includegraphics[width=0.85\linewidth]{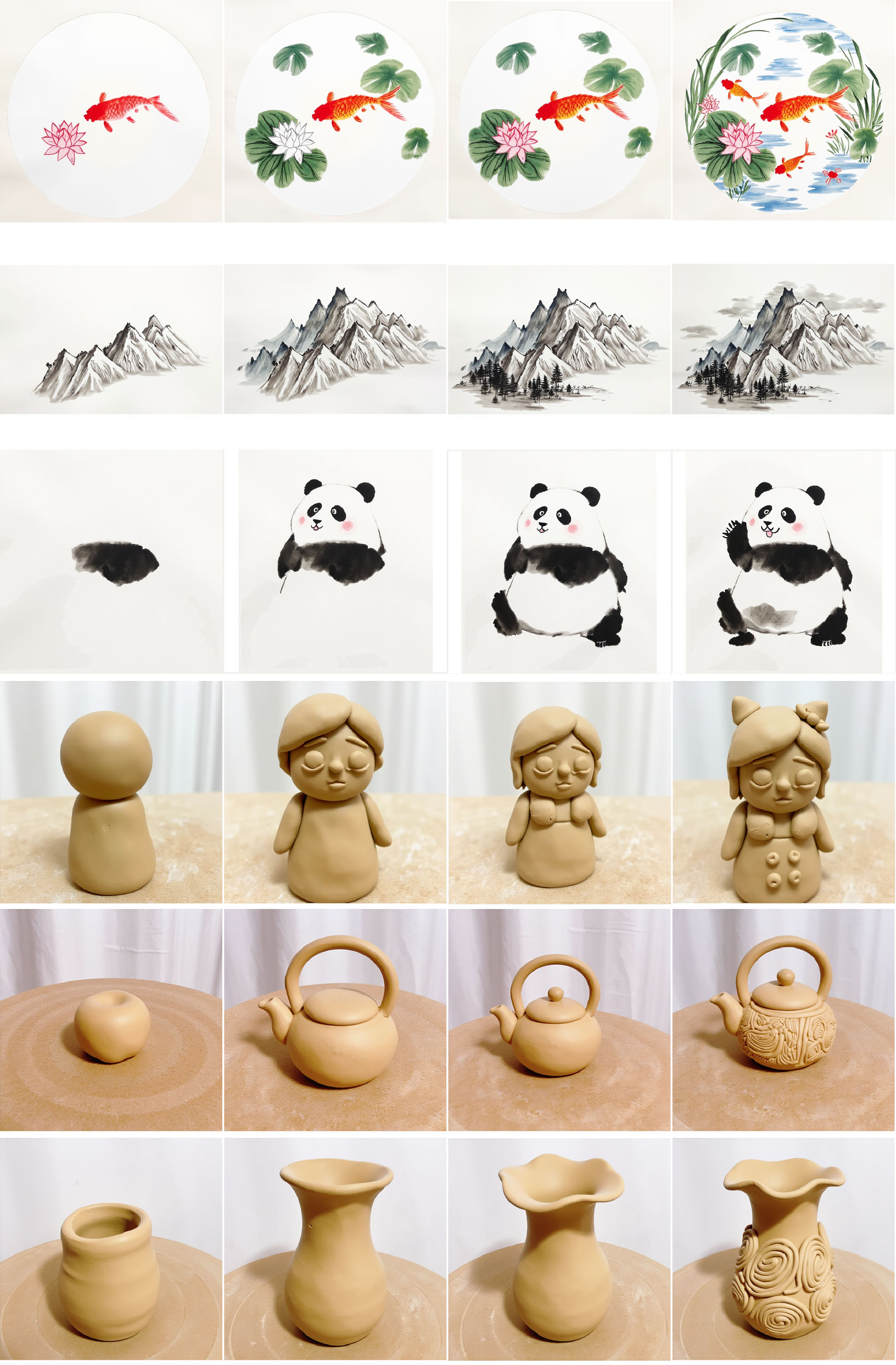} % Replace with your image file
    \caption{More generation results. From top to bottom, they are ink painting and clay sculpture.}
    \label{fig10}
\end{figure*}

\begin{figure*}[htp]
    \centering
    \includegraphics[width=0.85\linewidth]{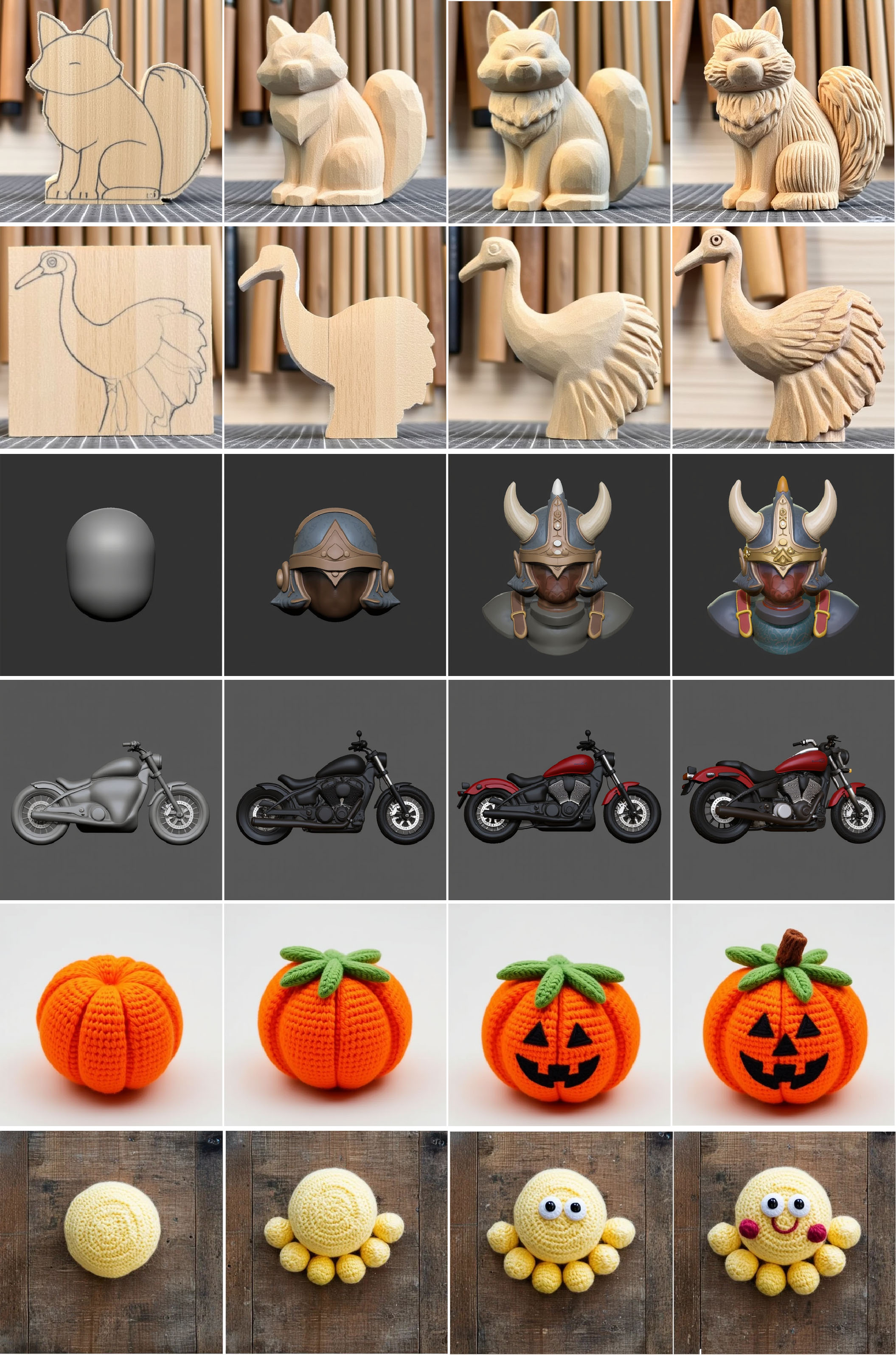} % Replace with your image file
    \caption{More generation results. From top to bottom, they are wood sculpure, Zbrush, and fabric toys.}
    \label{fig11}
\end{figure*}

\end{document}